\documentclass[runningheads]{llncs}

\usepackage{eccv}

\usepackage{eccvabbrv}

\usepackage{graphicx}
\usepackage{booktabs}

\usepackage[accsupp]{axessibility}  

\usepackage{hyperref}

\usepackage{orcidlink}
\usepackage{times}
\usepackage{helvet}
\usepackage{courier}
\usepackage{amsmath,amsfonts,amssymb}
\usepackage{algorithmic}
\usepackage{algorithm}
\usepackage{array}
\usepackage{textcomp}
\usepackage{stfloats}
\usepackage{url}
\usepackage{verbatim}
\usepackage{graphicx}
\usepackage{epsfig}
\usepackage{booktabs}
\usepackage{makecell}
\usepackage{multirow}
\usepackage{pifont}
\usepackage{tablefootnote}
\usepackage{enumitem}

\usepackage{amsmath}
\DeclareMathOperator\arctanh{arctanh}

\begin{document}

\title{Learning Multi-Manifold Embedding for Out-Of-Distribution Detection} 

\titlerunning{MMEL for OOD Detection}

\author{Jeng-Lin Li\inst{1}\orcidlink{0000-0002-9261-1524} \and
Ming-Ching Chang\inst{1,2}\orcidlink{0000-0001-9325-5341} \and
Wei-Chao Chen\inst{1}\orcidlink{0000-0002-6165-7706}}

\authorrunning{J-L.Li et al.}

\institute{Inventec Corporation, No.66, Hougang St., Shilin Dist., Taipei City 111059, Taiwan \\
\email{\{li.johncl, chang.ming-ching, chen.wei-chao\}@inventec.com}\\ \and
University at Albany, State University at New York, Albany, NY, 12222, USA\\
\email{mchang2@albany.edu}}

\maketitle

\begin{abstract}
  Detecting out-of-distribution (OOD) samples is crucial for trustworthy AI in real-world applications. Leveraging recent advances in representation learning and latent embeddings, Various scoring algorithms estimate distributions beyond the training data. However, a single embedding space falls short in characterizing in-distribution data and defending against diverse OOD conditions. This paper introduces a novel {\bf Multi-Manifold Embedding Learning (MMEL)} framework, optimizing hypersphere and hyperbolic spaces jointly for enhanced OOD detection. MMEL generates representative embeddings and employs a prototype-aware scoring function to differentiate OOD samples. It operates with very few OOD samples and requires no model retraining. Experiments on six open datasets demonstrate MMEL's significant reduction in FPR while maintaining a high AUC compared to state-of-the-art distance-based OOD detection methods. We analyze the effects of learning multiple manifolds and visualize OOD score distributions across datasets. Notably, enrolling ten OOD samples without retraining achieves comparable FPR and AUC to modern outlier exposure methods using 80 million outlier samples for model training. 
\end{abstract}

\section{Introduction}
\label{sec:intro}

In data-driven machine learning (ML), out-of-distribution (OOD) samples refer to unseen instances outside the distribution the ML models were trained on. Deploying artificial intelligence (AI) models often encounter OOD challenges due to domain shifts in test data compared to the original training data. This shift can cause trained models to be over-confident in incorrect decisions, leading to issues of trustworthiness and reliability.
Detecting OOD samples from in-distribution (ID) data is challenging due to the vast OOD sample space compared to the ID data. In standard image classification tasks, the training set is considered the ID dataset, while any images outside or significantly different from the training set are considered OOD samples.

Past research on OOD detection has predominantly focused on designing scoring functions based on predicting probabilities~\cite{shen2021towards,yang2021generalized}. The evolution of the scoring function includes approaches such as maximum softmax probability~\cite{hendrycks2016baseline} and energy-based scores~\cite{liu2020energy}. 
OOD detection performance can be enhanced through additional manipulation, such as perturbation and normalization~\cite{wei2022mitigating}. 
Beyond the use of logits, studies have explored manipulating network inputs and parameters. For instance, ODIN~\cite{liang2018enhancing} uses temperature scaling and introduces small perturbations to better distinguish ID and OOD images based on their softmax score distributions. Simple and effective approaches involve pruning and rescaling the network layers~\cite{djurisic2022extremely}. ViM~\cite{wang2022vim} introduces a virtual OOD class on top of known ID classes using both features and logits.

Recent studies have opened up a new avenue by enhancing the latent space of networks to establish better representations that capture relationships between samples. In these approaches, OOD detection is conducted by comparing the distance between embeddings. While using the Mahalanobis distance~\cite{lee2018simple} as a confident score is useful, its performance is limited by the unchanged network learning scheme. By employing supervised contrastive model training loss, the network performance can be enhanced. Examples like SSD~\cite{sehwag2020ssd} and KNN+~\cite{sun2022out} calculate scores based on Mahalanobis and non-parametric KNN distance, respectively. The CIDER framework~\cite{ming2023how} projects training data onto a hypersphere space, significantly improving OOD detection performance. However, prior research constraints the embedding to learn with a single manifold structure, leading to distorted representations that underrepresent part of the ID data.

In framing OOD detection as a representation learning problem, learning latent manifolds becomes crucial for enhancing the compactness of ID embeddings and the separability of OOD embeddings. 
Riemannian manifolds form powerful manifold spaces with curvature parameters signifying deviation from the Euclidean space, such as the {\em hypersphere} with positive curvature and {\em hyperbolic} spaces with negative curvature. Real-world data are mixed with spherical and hierarchical structures. Apart from the hypersphere characterizing class prototypes with the sphere centers~\cite{ming2023how}, the Imagenet dataset~\cite{ILSVRC15} demonstrates a natural hierarchical structure in the real world that can be represented in hyperbolic space.
Therefore, we project embeddings onto both hypersphere and hyperbolic spaces within a multi-manifold learning scheme.

\begin{figure*}[t]
\centerline{
  \includegraphics[width=\linewidth]{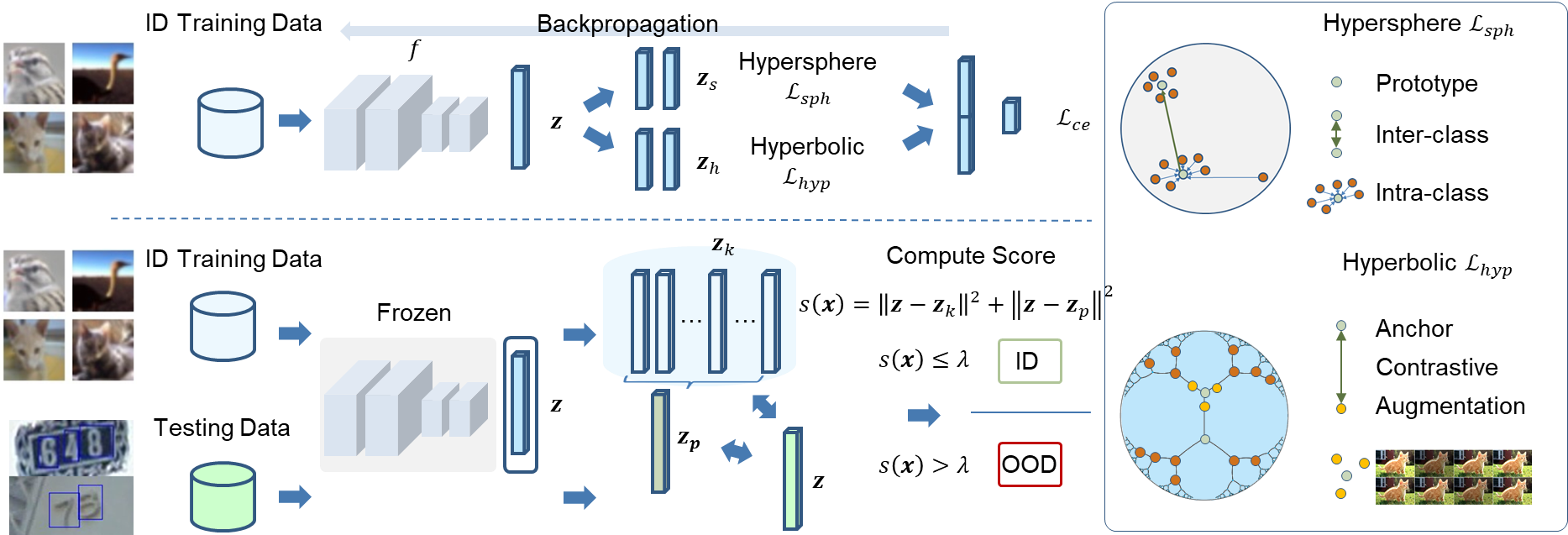} 
  \vspace{-3mm}
}
\caption{Overview of the proposed {\bf Multi-Manifold Embedding Learning (MMEL)} framework for OOD detection. The upper part indicates the network structure trained with the hypersphere and hyperbolic manifolds which are illustrated in the right box for details. The lower part indicates the OOD score computation in the inference phase.}
\label{fig:framework}
\vspace{-3mm}
\end{figure*}

We introduce a novel {\bf Multi-Manifold Embedding Learning (MMEL)} framework, which incorporates both positive and negative curvature manifolds to enhance latent representations for OOD samples; see Figure~\ref{fig:framework}. Through joint learning of multiple manifolds with multitask losses, our framework aims to diversify the embedding space, preserving latent manifolds with minimally distorted representation relations when encountering unknown OOD samples. Additionally, we design an enhanced KNN scoring function by considering ID cluster prototypes, providing a more nuanced characterization of testing samples relative to the training distribution. 

Based on the multi-manifold design, we raise the question of whether the framework can be more flexible to accomodate with new ID or OOD distributions as it encompasses multiple latent manifold structures. Therefore, we demonstrate a new usage scenario called {\em test-time OOD enrollment}, where a few OOD samples are collected during testing. In many practical applications, a few OOD samples are easy to collect, allowing the subsequent OOD detection for robust model deployment. We also examine the test-time novel ID class enrollment, including unseen ID classes for OOD detection.


Our MMEL framework outperforms other state-of-the-art (SoTA) OOD detection methods. Evaluation is performed using CIFAR-100 as the ID dataset and six other datasets as the OOD testing. MMEL achieves 10.26\% FPR$_{95}$, the false positive rate at 95\% true positive rate. 
Furthermore, our test-time OOD enrollment results demonstrate that enrolling as few as 10 OOD samples significantly reduces FPR$_{95}$ for OOD detection, eliminating the need of model retraining. This performance is comparable to the modern outlier exposure method~\cite{wu2023towards} trained on huge OOD datasets (over 80 million outlier samples). Additional experiments also demonstrate that ID space can be flexibly expanded by enrolling novel ID classes.

Our contributions are outlined as follows:
\begin{itemize}

\item We propose a new MMEL framework that incorporates both hypersphere and hyperbolic manifold representations, along with an advanced prototype-aware KNN scoring function for improved OOD detection. 

\item We show that MMEL outperforms the SoTA OOD detection methods on six open datasets using evaluation metrics reflecting low FPR$_{95}$ at high AUC.

\item We introduce a new test-time enrollment approach using as few as 10 OOD samples, showing comparable performance to other outlier exposure methods that require model retraining on huge OOD datasets.

\item We analyze the effects of learning multiple manifolds by visualizing the OOD score distributions. We also explore the benefits of the new enrollment approach showing the potential of MMEL in practical usage scenarios.
\end{itemize}

\section{Related Work and Preliminary}
\label{sec:related_work}

We review the hypersphere embedding, hyperbolic embedding, and multiple manifold learning within the context of a classification problem with the following notations. An input data sample, denoted as $\bf{x}\in \mathcal{X}$, undergoes processing by a model $f: \mathcal{X}\rightarrow \mathcal{Y}$ to predict a label $y\in \mathcal{Y}^{ID}$, where $ID$ denotes in-distribution. The training set contains $K$ classes, thus $\mathcal{Y}^{ID}=\{y_1, y_2, ..., y_K\}$. Model $f$ is trained using ID training data $\bf{x}$ sampled from the marginal distribution $P_\mathcal{X}^{ID}$ and produces the latent embedding $\bf{z}$.

During inference, our goal is to detect OOD samples from $P_\mathcal{X}^{OOD}$, where the corresponding OOD label space may extend beyond the $\mathcal{Y}^{ID}$ range. An estimator $g$ conducts OOD detection using a scoring function $S(\bf{z})$ and a threshold $\lambda$:
\begin{equation}
    g_\lambda(\bf{z}) = 
        \begin{cases}
            ID  &  \text{if}\ S(\bf{z})\leq \lambda, \\
            OOD   & \text{otherwise.}
        \end{cases}
\label{eq:ood_determination}
\end{equation}

\subsection{Hypersphere Embedding}

Hypersphere embedding stands out with remarkable success across various ML domains, including face verification~\cite{Liu2022SphereFaceR}, person re-identification~\cite{fan2019spherereid}, emotion recognition~\cite{li2022enroll}, and adversarial training~\cite{pang2020boosting}. 
It was first introduced in CIDER~\cite{ming2023how} as a learning method for OOD detection.
Hypersphere embedding learning methods often convert the standard cross-entropy loss into the angular space by eliminating the bias term. The resulting loss function comprises a radius term and an angular term. Since the radius term merely affects the scale, the attention focuses on optimizing the angular term. This loss function thus shapes the relationships of latent embeddings on a hypersphere. 
The latent embedding $\mathbf{u}$ is associated with an angle $\theta_y$ to the weight $\mathbf{W}$ and the corresponding label $y$, leading to the reformulated generalized loss function~\cite{Liu2022SphereFaceR}:
\begin{equation}
L_s = -\log 
  \left( 
  \frac{
    \exp \left( ||\mathbf{u}||\phi(\theta_y) \right)
  }{
    \exp \left(
      ||\mathbf{u}||\phi(\theta_y) 
    \right)
    +
    \sum_{i\neq y} \exp 
      \left(
        ||\mathbf{u}||\eta(\theta_i)
      \right)
  } 
  \right),
\nonumber
\end{equation}
where $\phi$ and $\eta$ are the angular functions for the target class and the other classes, respectively. By absorbing the negative sign in the previous equation and reorganizing the term for inter-class angular function, we can derive an angular margin $\Delta(\theta)=\eta(\theta_y)-\phi(\theta_y)$ to enlarge the inter-class distance and suppress the intra-class variability:
\vspace{-2mm}
\begin{equation}
\begin{split}
L_s = 
\log \left( 
  1 + \sum_{i\neq y}
  \exp \left( ||\mathbf{u}||
    \left( \eta(\theta_i)-\phi(\theta_y) 
    \right)
  \right)
\right) \\ 
\hspace{-2.5mm}= 
\log \left(
  1 + \sum_{i\neq y}
  \exp \left( ||\mathbf{u}|| 
    \left( \eta(\theta_i)-\eta(\theta_y)+\Delta(\theta) 
    \right)
  \right)
\right).
\end{split}
\label{eq:sphere_loss}
\end{equation}
Integrating with metric learning, cluster centers are denoted as {\em prototypes} to capture intra-class and inter-class relationships. Prototype-based losses leverage spherical properties through a mixture of von Mises-Fisher (vMF) distributions for OOD detection~\cite{ming2023how}.

\subsection{Hyperbolic Embedding} 

Hyperbolic embedding has demonstrated notable success in image recognition and person re-identification tasks~\cite{khrulkov2020hyperbolic}. Its effectiveness stems from the unique properties of the hyperbolic space, particularly its ability to handle hierarchical data structures. While hyperbolic embedding is commonly used in natural language and graph applications, its benefits extend to few-shot learning scenarios and image-related tasks, where diverse geometric structures in testing distributions necessitate the use of different curvatures~\cite{gao2021curvature}.
Recent advancements in hyperbolic embedding, particularly incorporating prototypes from metric learning, further enhance the discriminative power of the embedding space~\cite{ghadimi2021hyperbolic}.

\subsection{Multiple Manifold Learning}

Manifold learning aims to capture the latent structure of a dataset, facilitating the discovery of a low-dimensional space that offers a compact and effective representation. The key objective is to preserve the relationships between neighboring data points within the learned embedding space. Recent endeavors have expanded to the exploration of learning multiple manifolds, recognizing the manifold heterogeneity inherent in datasets~\cite{hettiarachchi2015multi}. These studies incorporate well-designed optimization strategies to ensure model convergence while learning multiple manifolds.

However, the aforementioned works primarily focus on exploring subspaces within Euclidean space. In contrast, another research direction delves into curved manifolds, defining mixed spaces that combine manifolds with different curvatures~\cite{gu2018learning}. This method has demonstrated impressive performance in benchmarks related to data reconstruction and word embeddings in natural language processing.

Despite the rapid advancements in hypersphere and hyperbolic embeddings, the exploration of hyperbolic space and joint spaces for OOD detection remains largely untapped. This presents an intriguing avenue for future research in this area.

\section{Method}
\label{sec:method}

Figure~\ref{fig:framework} overviews the proposed MMEL framework for OOD detection, including the training and inference steps.
The framework is constructed by integrating the hypersphere and hyperbolic branches through a multi-task joint loss optimization scheme. $\S$~\ref{ssec:multimanifold} presents the multi-manifold embedding learning. OOD scores are computed using the learned embeddings in $\S$~\ref{ssec:score_calculation}.
$\S$\ref{sec:test:enroll} presents our novel test-time enrollment approach for effective OOD detection without the need for model retraining.

We follow the standard OOD detection setup as follows:
(1) Train a model with ID training data and freeze model parameters.
(2) Run the model on test data.
(3) Calculate OOD scores and identify OOD samples using a threshold.

\subsection{Learning Multiple Manifold Embedding}
\label{ssec:multimanifold}

We next describe the hypersphere and hyperbolic embedding learning in the following section. Then, we describe the loss optimization using these learned embeddings. 

\vspace{-1mm}
\subsubsection{Learning hypersphere manifold}
\label{sssec:hypersphere}
\vspace{-1mm}

We use CIDER~\cite{ming2023how} to optimize compactness and disparity losses for a hypersphere manifold, represented by the von Mises-Fisher (vMF) distribution with a unit vector $\mathbf{z_s} \in \mathcal{R}_s^d$ in class $k$ and the class prototype $\boldsymbol{\mu}_k$:
\begin{equation}
    p_d(\mathbf{z_s};\boldsymbol{\mu}_k) = \tau \; \exp(\boldsymbol{\mu}_k\mathbf{z_s}/\tau),
\end{equation}
where $\tau$ is a temperature parameter, by default assigned as 0.1.
The probability of the embedding $\mathbf{z_s}$ assigned to class $k$ is:
\begin{equation}
    \mathcal{P}(y=k|\mathbf{z_s};\{\boldsymbol{\mu}_k, \tau\}) = \frac{\exp \left( \boldsymbol{\mu}_k\mathbf{z_s}/\tau \right)}{\sum_{j=1}^{K}{\exp \left( \boldsymbol{\mu}_j\mathbf{z_s}/\tau \right)}}.
\end{equation}
By taking negative log-likelihood, we obtain the compactness loss, which compels each sample to be close to the prototypes of its belonging class.
\begin{equation}
    \mathcal{L}_{com} = -\frac{1}{N}\sum_{j=1}^{K}\log\frac{\exp \left(
    \boldsymbol{\mu}_k\mathbf{z_s}/\tau \right)}{\sum_{j=1}^{K}{\exp \left( \boldsymbol{\mu}_j\mathbf{z_s}/\tau \right)}}.
\label{eq:comp_loss}
\end{equation}
The disparity loss encourages a large angular margin among class prototypes:
\begin{equation}
    \mathcal{L}_{dis} = \frac{1}{K}\sum_{i=1}^{K}\log\frac{1}{K-1}\sum_{j=1}^{K}\mathbf{1}_{ji}
    \exp{ 
      \left( \boldsymbol{\mu}_i\boldsymbol{\mu}_j/\tau
      \right)
    },
\label{eq:disparity_loss}
\end{equation}  
where indication function
$\mathbf{1}_{ji} =
    \begin{cases}
      1 & \text{if $j\neq i$},\\
      0 & \text{otherwise.}
    \end{cases}$
The loss function for the hypersphere branch is given by $\mathcal{L}_{sph} = \mathcal{L}_{com} + \mathcal{L}_{dis}$. These two losses jointly shape the clusters on the hypersphere, ensuring intra-class compactness and inter-class disparity for ID data. As a result, OOD data are less likely to reside in the space near ID prototypes.

\subsubsection{Learning hyperbolic manifold}
\label{sssec:hyperbolic}



We are the first work introducing hyperbolic manifold for OOD detection. An $d$-dimensional hyperbolic space $H^d$ is a collection of $d$-dimensional Riemannian manifolds with constant negative curvature~\cite{khrulkov2020hyperbolic,9880306}, where the curvature indicates the deviation from Euclidean space. Among various models formulated for isomorphic transformation between hyperbolic spaces, the Poincaré Ball is represented as $\mathbb{M}^d_c$ with curvature $c$. 
Based on the embedding $\mathbf{u}$, the manifold is defined as $\mathbb{M}^d=\{\mathbf{u}\in \mathbb{R}^d: c||\mathbf{u}||<1\}$, and the Riemannian metric tensor $g^\mathbb{M}(\mathbf{u})$ is expressed as $(\lambda^c_\mathbf{u})^2g^E = \left(\frac{2}{1-c||\mathbf{u}||^2} \right)^2\mathbf{I}$, where $\lambda = \frac{2}{1-c||\mathbf{u}||^2}$ is a conformal factor, and $g^E=\mathbf{I}$ is the Euclidean metric tensor.

In the manifold, we need operations from Mobius gyrovector space, including Mobius addition $\oplus_c$ and scalar multiplication $\otimes_c$ for vectors ($\mathbf{u}$ and $\mathbf{v}$) with the scalar $w$.
\begin{equation}
    \mathbf{u}\oplus_c\mathbf{v} = \frac{(1+2c<\mathbf{u},\mathbf{v}>+c||\mathbf{v}||^2)\mathbf{u}+(1-c||\mathbf{u}||^2)\mathbf{v}}{1+2c<\mathbf{u},\mathbf{v}>+c^2||\mathbf{u}||^2||\mathbf{v}||^2},
\nonumber
\end{equation}
\begin{equation}
    w \otimes_c \mathbf{u} = \frac{1}{\sqrt{c}}\tanh{ \left(w\cdot \arctanh(\sqrt{c}||\mathbf{u}||) \right)}\frac{\mathbf{u}}{||\mathbf{u}||},
\end{equation}
The geodesic distance between two points $\mathbf{u}$ and $\mathbf{v}$ is calculated by:
\begin{equation}
    D(\mathbf{u}, \mathbf{v}) = \frac{2}{\sqrt{c}}\arctanh \left(\sqrt{c}||-\mathbf{u} \oplus_c \mathbf{v}|| \right).
\label{eq:hypb_dist}
\end{equation}
As the curvature $c$ approaches $0$, the distance converges to $2||\mathbf{u}-\mathbf{v}||$, which reduces to Euclidean distance. 

We utilize an {\em exponential map} to transform a vector to the tangent space on the Poincaré ball. The embedding vector $\mathbf{v}$ generated by a backbone network, is transformed into hyperbolic embedding using the exponential map ${\cal E}^c (\mathbf{v}) = \tanh{ \left(\sqrt{c}||\mathbf{v}||\right)\frac{\mathbf{v}}{\sqrt{c}||\mathbf{v}||}}$.
Subsequently, we apply {\em hyperbolic averaging} to multiple hyperbolic embeddings via Einstein midpoint. The embedding from the Poincaré ball $\mathbb{D}_c^d$ 
 can be projected to the Klein model $\mathbb{K}_c^d$, allowing for a simpler average calculation in the  Klein coordinate:
\begin{equation}
\mathbf{u_\mathbb{K}} = \frac{2\mathbf{u}_\mathbb{D}}{1 + c \; ||\mathbf{u}_\mathbb{D}||^2}, \;\;
\overline{\mathbf{u}_\mathbb{K}}=\frac{\sum_{i=1}^{m}r_i\mathbf{u}_{\mathbb{K},i}}{\sum_{i=1}^{m}r_i},
\end{equation}
where $r_i$ is the Lorentz factor. After deriving the average embedding in the Klein coordinate, we transform the space back to the Poincaré ball:
\begin{equation}
\overline{\mathbf{u}_\mathbb{D}}=\frac{\overline{\mathbf{u}_\mathbb{K}}}{1+\sqrt{1-c \; ||\overline{\mathbf{u}_\mathbb{K}}||^2}}.
\end{equation}

Using the operations available in the hyperbolic space, we project the latent embedding with a hyperbolic head to obtain the embedding $\mathbf{z_h}$ on the Poincaré ball. Creating an augmented set $\mathcal{A}$ from $\mathcal{X}$ to form a full set $\mathcal{I}=\mathcal{A}\cup \mathcal{X}$, we calculate the supervised contrastive loss on the positive sample $p(i)$ of the $i\in \mathcal{I}$ in contrast to other augmented samples $a\in \mathcal{A}$. We denote the embeddings of positive samples and augmented samples as $\mathbf{z_h}_p$ and $\mathbf{z_h}_a$.
The supervised hyperbolic contrastive loss can thus be formulated as $\mathcal{L}_{hypb} = $
\begin{equation}
-\sum_{i\in \mathcal{I}}\frac{1}{|P(i)|}\sum_{p\in P(i)}\log\frac{\exp \left( -D(\mathbf{z_h}_i, \mathbf{z_h}_p)/\tau \right) }{\sum_{a\in \mathcal{A}}{\exp \left( -D(\mathbf{z_h}_i, \mathbf{z_h}_a)/\tau \right)}}.
\nonumber    
\end{equation}

\subsubsection{Loss optimization}
\label{sssec:optimization}

The overall loss encompasses the hypersphere and hyperbolic losses along with a cross-entropy loss $\mathcal{L}_{ce}$ to optimize for ID classification accuracy: 
\begin{equation}
    \mathcal{L} = \mathcal{L}_{sph} + \mathcal{L}_{hypb} + \mathcal{L}_{ce}.
\end{equation}
The curvature parameter $c$ in Eq.~\eqref{eq:hypb_dist} is typically treated as a hyperparameter, and we choose its value by referring to the Gromov measurement~\cite{khrulkov2020hyperbolic}. 
To enhance stability during learning, we employ an empirically found feature clipping technique~\cite{9880306}, which involves truncating a Euclidean space sample point $\mathbf{x}$ as the clipped feature $\mathbf{x'}=\min\{1, \frac{r}{||\mathbf{x}||}\}\cdot \mathbf{x}$, with an effective radius $r$ for the Poincaré ball. This helps avoid gradient vanishing in complex manifold learning and regularizes the points close to the ball boundary. 


\subsection{OOD Score Calculation}
\label{ssec:score_calculation}

Upon obtaining a trained network $f$ within the MMEL framework, we extract the penultimate layer output as an L2 normalized embedding $\mathbf{z}$ of the sample $\mathbf{x}$ to compute its OOD score. 
To distinguish between OOD and ID samples, we measure the embedding distance between each input sample and specified training ID samples acting as reference anchors. 

Initially, We utilize the $k$-th nearest neighbor (KNN) as a reference embedding $\mathbf{z}_k$ for distance computation. The resulting OOD score from this $k$-th nearest neighbor is denoted as $S_k(\mathbf{z})$. Additionally, we calculate the distance to the nearest $p$ training cluster centers, deriving the average of these $p$ distance values as the score $S_p(\mathbf{z})$.

\medskip
\noindent
{\bf Prototype-aware KNN (PKNN) OOD score calculation.}
To ensure robust OOD score calculation, we consider multiple anchors, including the $k$-th training samples and the $p$ nearest cluster centers in the calculation using: 
\begin{equation}
S(\mathbf{z}) = S_k(\mathbf{z}) + S_p(\mathbf{z}) = ||\mathbf{z}-\mathbf{z}_k||^2 + \frac{1}{p}\sum_{p}{||\mathbf{z}-\boldsymbol{\mu}_p||^2},
\nonumber
\end{equation}
where the OOD score is obtained based on the L2 distance.
This explicitly improves OOD estimation robustness, in contrast to the previous works~\cite{sun2022out,ming2023how}, where only the $k$-th training samples are used for OOD score calculation. Finally, the OOD detection is performed using Eq.~\eqref{eq:ood_determination}. 

\subsection{Test-time OOD Sample Enrollment} 
\label{sec:test:enroll}

With ability to capture multi-manifold structures, We explore a novel usage scenario to further enhance OOD detection performance. In many real-world applications,] continuous occurrences of OOD samples sharing underlying characteristics may be frequently encountered. Suppose very few OOD samples can be collected beforehand, we can incorporate the knowledge of these potential OOD samples into the OOD detection framework. We term this approach as {\bf OOD sample enrollment}. 

Specifically, in the test time of OOD detection, we compute the average embedding vector of the obtained $N_e$ OOD samples as an enrolled prototype $\mathbf{z}_e$. Our assumption is that the test samples are likely to be OOD samples if they are close to these enrolled OOD prototypes. Utilizing the OOD scoring function, we calculate the L2 distance between the test sample embedding $\mathbf{z}$ and the enrolled prototype $\mathbf{z}_e$ as an additional negative OOD score $-S'(\mathbf{z})=-||\mathbf{z}-\mathbf{z}_e||^2$, resulting in the final OOD score as $S(\mathbf{z})-S'(\mathbf{z})$.
Our proposed OOD enrollment framework is flexible and can be quickly applied in various OOD scenarios without requiring model retraining. 

\begin{table}[t]
\caption{Evaluation of OOD detection using {\em CIFAR-10}, {\em CIFAR-100}, and {\em ImageNet-100} as ID samples and the other six datasets as OOD samples. We show the averaged FPR$_{95}$ and AUC scores across the six tests. {\em MMEL} achieves the best averaged FPR$_{95}$ and AUC. 
\vspace{-2mm}
}
\label{tab:cifar10_100_result}
\centerline{
\setlength{\tabcolsep}{0.8mm}
\begin{tabular}{l|cc|cc|cc}
\toprule
ID Dataset              & \multicolumn{2}{c}{CIFAR-10} & \multicolumn{2}{c}{CIFAR-100} & \multicolumn{2}{c}{ImageNet-100}\\
\midrule
Method  & 
FPR$_{95}${\footnotesize $\downarrow$} & 
AUC{\footnotesize $\uparrow$} & 
FPR$_{95}${\footnotesize $\downarrow$} & 
AUC{\footnotesize $\uparrow$} & 
FPR$_{95}${\footnotesize $\downarrow$} & 
AUC{\footnotesize $\uparrow$} \\
\midrule
MaxSoftmax  & 38.97        & 90.44        & 88.78       & 58.99   & 72.35 & 74.61 \\
Mahalanobis  & 25.30        & 93.69       & 72.21        & 74.22  & 54.21 & 83.80 \\
ODIN        & 40.17        & 91.16        & 81.57        & 68.05  & 82.65 &  65.38 \\
Energy      & 38.64        & 91.92        & 83.97        & 63.75  & 57.64 &  82.25 \\
Entropy     & 32.18        & 91.59        & 88.62        & 60.42   & 68.60 & 78.36 \\
ViM         & 29.17        & 92.98        & 75.94        & 73.34   & 54.63 & 77.53      \\
KLMatching  & 65.49        & 87.99        & 94.57        & 44.52  & 86.14 & 66.55  \\
MaxLogit     & 38.72        & 76.63       & 84.35        & 63.45   & 58.95 & 81.13 \\
GODIN        & 26.14        & 93.88       & 72.76        & 86.57  & 88.37 & 71.43 \\ 
DICE         &  20.83       & 95.24       & 49.72        & 87.23  & 35.76 & 90.66 \\
\midrule
SSD         & 27.45        & 96.08        & 70.98        & 84.94  & 32.99 & 94.11\\
KNN+        & 14.95       &  97.10        & 65.47        & 85.07  & 33.04 & 93.57 \\
CIDER       & 16.67        & 97.02        & 52.35        & 86.72  & 25.90 & 94.46 \\
\midrule
MMEL        & \textbf{14.15} & \textbf{97.52}  & \textbf{42.61} & \textbf{89.62} & \textbf{24.05} & \textbf{94.96} \\
\bottomrule
\end{tabular}
}
\vspace{-3mm}
\end{table}

\begin{table*}[t]
\caption{Comparison of {\em MMEL} to other manifold learning methods and ablation results for OOD detection on {\em CIFAR-100}.
\vspace{-2mm}
}
\label{tab:manifold_result}
\centerline{
\setlength{\tabcolsep}{-0.0mm}
\begin{tabular}{l|cc|cc|cc|cc|cc|cc|cc}
\toprule
CIFAR100       & \multicolumn{2}{c}{SVHN} & \multicolumn{2}{c}{Places365} & \multicolumn{2}{c}{LSUN} & \multicolumn{2}{c}{LSUN-R} & \multicolumn{2}{c}{iSUN} & \multicolumn{2}{c}{Texture} & \multicolumn{2}{c}{Average} \\
\midrule
Method & FPR$_{95}$ & AUC & FPR$_{95}$ & AUC & FPR$_{95}$ & AUC & FPR$_{95}$ & AUC & FPR$_{95}$ & AUC & FPR$_{95}$ & AUC & FPR$_{95}$ & AUC \\
\midrule
SF2            & 37.07       & 93.09      & 79.13         & 76.78         & 49.64       & 87.11      & 72.31        & 82.88        & 69.25       & 83.38      & 47.62        & 89.98        & 59.17        & 85.54        \\
SPH            & 15.41       & 96.00      & 86.85         & 65.47         & 66.50        & 81.51      & 75.26        & 82.78        & 76.78       & 81.22      & 57.18        & 85.60         & 63.00        & 82.10         \\
SFRN           & 36.99       & 92.34      & 79.30         & 78.23         & 53.25       & 86.18      & 66.62        & 84.73       & 68.41       &  83.93     & 57.04        & 87.22        & 60.27        & 85.44        \\
SFRH           & 58.23       & 88.98      & 82.64         & 73.91         & 83.24       & 76.61      & 79.20         & 80.59        & 81.82       & 79.25      & 73.21        & 84.40         & 76.39        & 80.62        \\
SFRS           & 48.23       & 90.80       & 84.98         & 73.22         & 73.66       & 78.65      & 84.95        & 77.44        & 84.28       & 77.46      & 69.17        & 84.60        & 74.21        & 80.36        \\
CIDER    & 15.28     &  96.81     & 79.98         & 74.15         & 26.40        & 93.01      & 69.73        &  84.05       & 73.29       & 82.52      & 49.40       & 89.77        & 52.35        & 86.72        \\
Hyperbolic     & 47.06       & 90.83      & 79.54         & 78.34        & 79.54       & 78.34      & 82.93        & 78.56        & 83.59       & 77.12      & 83.59        & 77.12        & 65.44        & 84.20        \\
\midrule
MMEL$_{\text{KNN}}$  & 17.14 & 95.61 & 76.71 & 77.52 & 24.42 & 94.28 & 62.18 & 82.97 & 61.46 & 82.84 & 34.82 & 90.95 & 46.12 & 87.36 \\
MMEL$_{\text{Maha}}$ & \textbf{12.28} & \textbf{97.12} & 77.23 & \textbf{79.09} & 21.20  & \textbf{96.17} & 77.35 & 80.57 & 80.61 & 78.79 & 63.99 & 84.26 & 55.44 & 86.00    \\
MMEL$_{\text{PKNN}}$ & 13.98 & 96.83 & \textbf{75.05} & 78.94 & \textbf{20.13} & 95.88 & \textbf{58.09} & \textbf{86.33} & \textbf{58.09} & \textbf{86.33} & \textbf{31.44} & \textbf{93.33} & \textbf{42.61} & \textbf{89.62} \\
\bottomrule
\end{tabular}
}
\vspace{-4mm}
\end{table*}

\section{Experiments}
\label{sec:experiment}

\noindent
{\bf Dataset.}
We use CIFAR-10 and CIFAR-100~\cite{krizhevsky2009learning} as the ID dataset and examine the performance on six other datasets that are treated as OOD: SVHN~\cite{netzer2011reading}, Place365~\cite{7968387}, LSUN~\cite{yu2015lsun}, LSUN-Resize~\cite{yu2015lsun}, iSUN~\cite{xu2015turkergaze}, and Textures~\cite{6909856}.  
In another experiment, we follow the setup in \cite{ming2023how} and adopt the ImageNet-100 dataset as ID data which subsampled 100 classes from ImageNet~\cite{ILSVRC15}. Here, the other datasets regarded as OOD include SUN~\cite{xiao2010sun}, Place365~\cite{7968387}, Textures~\cite{6909856}, and iNaturalist~\cite{van2018inaturalist}.

\medskip
\noindent
{\bf Evaluation metric.}
All methods are evaluated using two common OOD detection metrics:
(1) FPR$_{95}$: False positive rate at true positive rate of 95\%.
(2) AUC: Area under the Receiver Operating Characteristic curve.


\subsection{Out-of-distribution Detection Accuracy}
\label{sssec:ood_accuracy}

We use the ResNet-18 backbone network for CIFAR-10 and ResNet-34 for CIFAR-100 to assess OOD performance. To gauge generalization to the ImageNet-100 dataset, we fine-tune the model trained on CIFAR-100 for experiments. We follow the parameter setting of CIDER~\cite{ming2023how} to ensure comparable results.  

The model is optimized via stochastic gradient descent (SGD) with momentum 0.9, weight decay of $10^{-4}$, and an initial learning rate of 0.5. Batch size and total epochs are fixed at 512 and 500, respectively. The intermediate layer comprises a 128-dimensional projection head. For ImageNet-100 fine-tuning, we employ a learning rate of 0.01 for 10 epochs. The curvature $c$ of hyperbolic geometry is chosen to be $0.01$. PKNN is implemented using Faiss-GPU~\cite{johnson2019billion} with $k=300$ and $p=3$.

We compare MMEL against 10 popular OOD detection methods, including MaxSoftmax~\cite{hendrycks2016baseline}, Mahalanobis~\cite{lee2018simple}, ODIN~\cite{liang2018enhancing}, Energy~\cite{liu2020energy}, Entropy~\cite{chan2021entropy}, ViM~\cite{wang2022vim}, KLMatching~\cite{basart2022scaling}, MaxLogits~\cite{basart2022scaling}, GODIN~\cite{hsu2020generalized}, and DICE~\cite{sun2022dice}.
We also compare with three embedding-based methods, namely SSD~\cite{sehwag2020ssd}, KNN+~\cite{sun2022out}, and CIDER~\cite{ming2023how}.

Table~\ref{tab:cifar10_100_result} shows that our MMEL framework outperforms other OOD detection approaches in the average FPR$_{95}$ across six datasets, specifically, 14.15\% and 42.61\% FPR$_{95}$ when using CIFAR-10 and CIFAR-100 as ID datasets, respectively.
The margin of FPR$_{95}$ becomes obvious with a larger ID class numbers (CIFAR-100), showcasing MMEL improvements on reducing FPR$_{95}$ by 9.74\% and 7.11\% over the best-performed distance-based and score-based methods from other OOD detection studies, respectively. 

The last column of Table~\ref{tab:cifar10_100_result} shows that MMEL excels in large-scale OOD detection on ImageNet-100, achieving 24.05\% FPR$_{95}$ and 94.96\% AUC. The fine-tuned results on the hundred classes closely align with reported outcomes for the entire ImageNet in other comparative methods~\cite{sun2022dice,sun2022out}. Results indicate that using CIFAR-100 as the ID dataset is more challenging due to the smaller training set compared to ImageNet-100. Notably, MMEL greatly outperforms the other methods on CIFAR-100.


\subsection{In-distribution Classification Accuracy} 

It is a trade-off between the OOD detection performance and the underlying model's classification accuracy, which requires dedicated balance in practice.
We experimented on CIFAR-100 to examine the underlying model's classification accuracy of MMEL and compare it with CIDER. Our result shows that CIDER achieves classification accuracy of 75.35\% with an OOD FPR$_{95}$ of 52.35\%. MMEL outperforms CIDER with 75.99\% classification accuracy, and also with a lower 46.12\% OOD FPR$_{95}$. The ID accuracy of CIFAR-10 is 94.53\%, 94.59\%, and 94.61\% for SSD, CIDER, and MMEL. Score-based algorithms ({\em e.g.}, GODIN, DICE) obtain equal accuracy (94.52\%). MMEL outperforms these score-based algorithms (74.60\%) in CIFAR-100, without showing a tradeoff between ID and OOD performance.  
This outcome affirms that incorporating additional manifolds in MMEL improves OOD detection; meanwhile, it does not compromise ID classification accuracy. 




\begin{table}[t]
\addtolength{\tabcolsep}{-3pt}
\caption{The ID accuracy (ID), averaged FPR$_{95}$, and AUC using {\em CIFAR-100} as ID samples. Different backbone network architectures are applied with CIDER and MMEL. 
\vspace{-4mm}
}
\label{tab:add_comparison}
\centerline{
\setlength{\tabcolsep}{0.5mm}
\renewcommand{\arraystretch}{0.9}
\begin{tabular}{l|ccc|ccc|ccc|ccc}
\toprule
& \multicolumn{3}{c}{ResNet34}  & \multicolumn{3}{c}{ResNet50}  & \multicolumn{3}{c}{DenseNet100} & \multicolumn{3}{c}{ViT} \\
\midrule
 &   ID &  FPR$_{95}${\footnotesize $\downarrow$} & AUC{\footnotesize $\uparrow$}  &   ID &  FPR$_{95}${\footnotesize $\downarrow$} & AUC{\footnotesize $\uparrow$}  &   ID  & FPR$_{95}${\footnotesize $\downarrow$} & AUC{\footnotesize $\uparrow$}   &   ID  & FPR$_{95}${\footnotesize $\downarrow$} & AUC{\footnotesize $\uparrow$}   \\
\midrule
CIDER   & 75.35 & 52.35 & 86.72 & 75.41     &  50.23   & 86.47  & 76.00   & 48.93     & 87.76 & 73.27 &  50.39  & 88.24  \\
MMEL  & \textbf{75.99} &  \textbf{42.61}  & \textbf{89.62}  & \textbf{76.02}  & \textbf{41.12}     & \textbf{90.01}   & \textbf{76.28}      & \textbf{41.83}  & \textbf{90.09}  & \textbf{74.66} & \textbf{46.17}      & \textbf{89.98} \\
\bottomrule
\end{tabular}
}
\vspace{-6mm}
\end{table}

\subsection{Ablation Studies of Different Manifolds}
\label{ssec:exp_manifold}

In this ablation study, we examine single-manifold embedding learning approaches and various scoring functions. In addition, we broaden the comparison by incorporating other renowned hypersphere manifold learning methods prevalent in the face and speaker verification domains. Specifically, we consider recent hypersphere embedding approaches including SphereFace2~\cite{wen2022sphereface}, SphereFace-R~\cite{Liu2022SphereFaceR}, and Spherized layer~\cite{kim2022spherization}. In Table~\ref{tab:manifold_result}, we denote SphereFace2 and Spherized layer as SF2 and SPH, respectively. SphereFace-R encompasses a number of different loss function designs, which are denoted as SFRH, SFRN, and SFRS, respectively. Most of these approaches originate from the face or speaker verification tasks. Our comparison includes them to assess the performance of alternative hyperspherical projections when using CIFAR-10 as the ID dataset. 

Table~\ref{tab:manifold_result} also reports the performance of (1) CIDER with hypersphere and (2) hyperbolic embedding learning described in $\S$\ref{sssec:hyperbolic}. In the hypersphere space, CIDER generally obtains the best performance, while SF2 outperforms CIDER on the iSUN and Texture datasets with FPR$_{95}$ of 69.25\% and 47.62\%, respectively. Solely relying on hyperbolic embedding proves less advantageous for OOD detection compared to CIDER. 
In contrast, MMEL achieves the best performance by jointly modeling two manifold spaces. It is worth highlighting that CIDER, with its compactness loss in Eq.~\eqref{eq:comp_loss} and disparity loss in Eq.~\eqref{eq:disparity_loss}, explicitly optimizes the relationship between each sample and the prototypes. This optimization leads to improvements over other sphere projection methods that emphasize only the angular margin in Eq.~\eqref{eq:sphere_loss}.

The last three rows of Table~\ref{tab:manifold_result} assess the use of various scoring functions in MMEL, including KNN, Mahalanobis (Maha), and our proposed PKNN in $\S$\ref{ssec:score_calculation}. Notably, KNN achieves 46.12\% FPR$_{95}$ and 87.36\% AUC, which outperforms other algorithms. However, PKNN reduces FPR$_{95}$ by 3.49\% and increases AUC by 2.26\% when compared with KNN. 
While Mahalanobis exhibits a less favorable average performance across the six datasets, it excels with a remarkable 12.28\% FPR$_{95}$ and 97.12\% AUC on the SVHN dataset. The design of Mahalanobis focuses solely on the distance to the cluster center, while KNN only considers the specified ID anchor sample. Mahalanobis proves effective for distant and easier OOD datasets like SVHN but falls short for other more challenging datasets. In contrast, our proposed PKNN combines the strengths of KNN with cluster prototypes, proving more advantageous across diverse OOD datasets.

\noindent\textbf{Different network architectures}: We investigate ResNet50, DenseNet100, and ViT as alternative backbone network architectures using the CIFAR-100 dataset as ID data. 
The slightly favorable ID accuracy with MMEL over CIDER can be observed in ResNet50 and DenseNet100 while the ViT tends to overfit, leading to accuracy degradation.  
The ability of the backbone network is proportional to OOD detection ability. Our proposed MMEL outperforms CIDER across these architectures. 

\begin{figure*}[t]
\centerline{
  {\footnotesize (a)}
  \includegraphics[width=0.48\linewidth,height=0.25\linewidth]{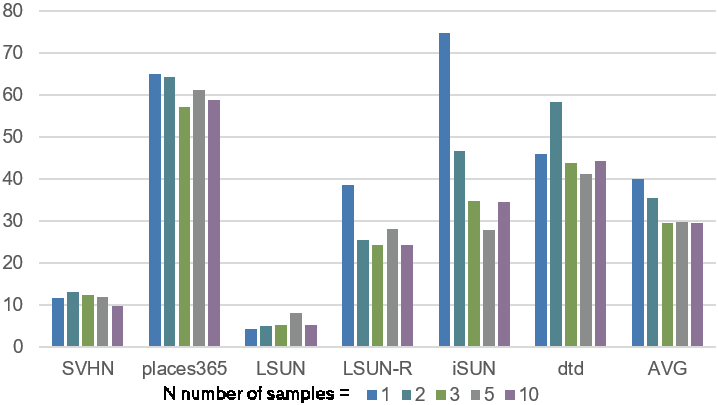}
  {\footnotesize (b)}
  \includegraphics[width=0.48\linewidth,height=0.25\linewidth]{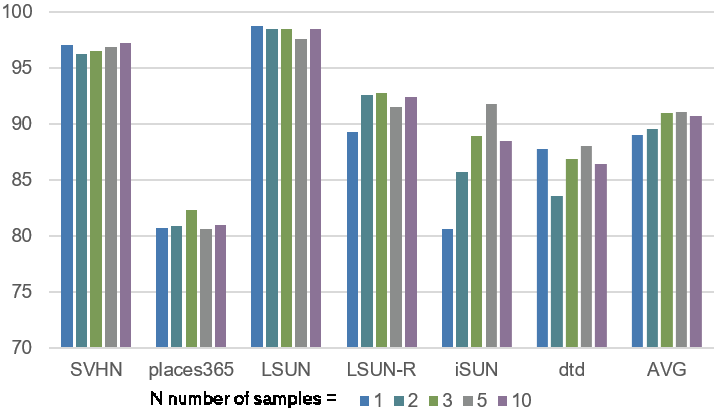}
\vspace{-2mm}
}
\caption{Results of (a) FPR$_{95}$ and (b) AUC percentage (\%) using $N_e$ OOD samples as a negative anchor, where each bar denotes the score using $N$ numbers of samples. {\em  CIFAR-100} is used as the ID samples and test on the six OOD datasets.
}
\label{fig:negative_anchor}
\vspace{-3mm}
\end{figure*}

\subsection{Visualization of OOD Scores}
\label{ssec:visualization}

\begin{figure}[t]
\centerline{
  \includegraphics[width=\linewidth,height=0.7\linewidth]{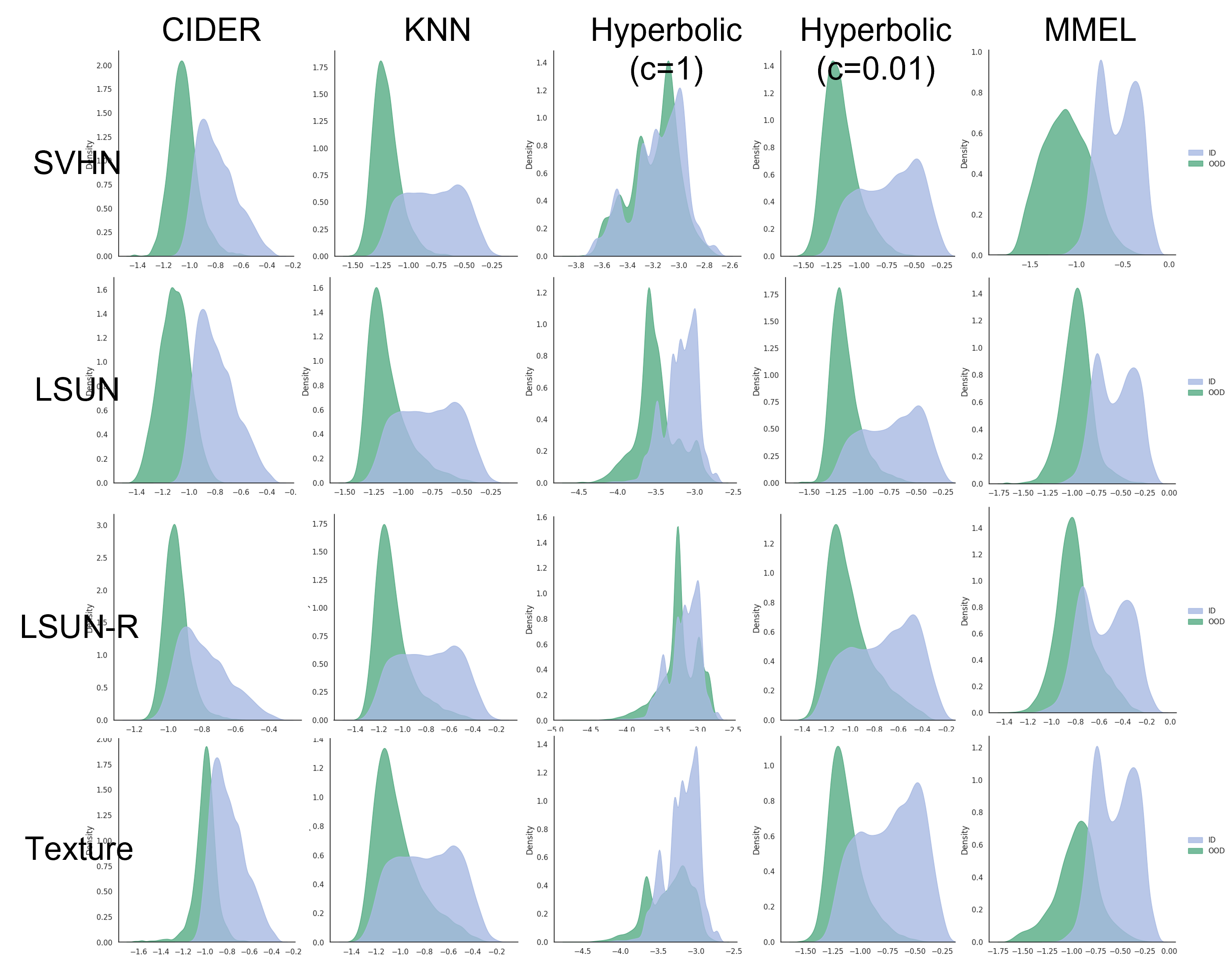}
  \vspace{-1mm}
}
\caption{Histogram visualization of OOD scores for ID samples in blue and OOD samples in green.}
\label{fig:score}
\vspace{-1mm}
\end{figure}

The OOD score plays a pivotal role in determining OOD detection outcomes and provides insights into the underlying distribution of ID and OOD data. To visually present these distributions, Figure~\ref{fig:score} plots the histogram of OOD detection scores for all test samples in each case, coloring ID samples in blue and OOD samples in green. Each row shows plots for each dataset, and each column shows plots for each algorithm. Greater separability in scores between ID and OOD histograms suggests better OOD detection performance.

Notably, with our MMEL, the ID histograms exhibit two distinct peaks in these plots, while the OOD histograms exhibit only one peak. In contrast, CIDER with hypersphere embedding yields a single peak, while hyperbolic embedding yields multiple peaks. MMEL allows more flexible learning to capture diverse latent space patterns.

\subsection{Evaluation of Test-Time Sample Enrollment}
\label{ssec:exp_enroll}

We evaluate the OOD enrollment approach (in $\S$\ref{sec:test:enroll}) for various OOD detection scenarios. 
$\S$\ref{sssec:improve_enroll} shows enhanced results using very few enrolled OOD samples compared to the best MMEL results in $\S$\ref{sssec:ood_accuracy}. 
$\S$\ref{sssec:enroll_class} further investigates a scenario where the ID dataset is expanded with additional classes. We discuss the different scenarios for enrolling the new classes in either the ID or OOD set for evaluation.


\vspace{-2mm}
\subsubsection{Improvement from enrolling a few OOD samples}
\label{sssec:improve_enroll}
\vspace{-1mm}

Figure~\ref{fig:negative_anchor} shows the results with varying values of $N_e$ within the set \{1, 2, 3, 5, 10\} using CIFAR-100 as ID data. The averaged FPR$_{95}$ across six OOD datasets decreases as $N_e$ increases, plateauing after $N_e$ suppresses three. Notably, incorporating enrolled samples leads to a 16.68\% reduction in FPR$_{95}$ compared to the scenario without any enrolled samples. Although the AUC metric sees only a marginal improvement, it follows a similar trend. It's worth noting that the decline of FPR$_{95}$ is not homogeneous across all OOD datasets. For instance, the iSUN dataset sees a substantial drop from 61.46\% to 34.57\% of FPR$_{95}$, with the enrollment of ten OOD samples. On the other hand, datasets like SVHN and LSUN, which already exhibit low FPR$_{95}$ with our MMEL framework, show limited benefits from the enrollment approach. These findings indicate that a significant improvement is achievable in FPR$_{95}$ by enrolling a small number of known OOD samples. Additionally, the results highlight the advantages of distance-based embedding learning methods, facilitating straightforward prototype estimation with new anchor samples.

We next compare experimental results against the outlier exposure method~\cite{wu2023towards}, which leverages an auxiliary dataset to learn the OOD space. Note that the selection strategy for this auxiliary dataset requires further investigation, and the inclusion of outlier training may compromise ID accuracy. 

Table~\ref{tab:outlier exposure} shows our OOD detection results comparing with ICE~\cite{wu2023towards}, the state-of-the-art outlier exposure method. ICE achieves 34.96\% FPR$_{95}$ and 90.90\% AUC through training on an 80-million auxiliary OOD dataset. In comparison, our MMEL, utilizing only 10 samples for enrollment, yields comparable results of 30.48\% FPR$_{95}$ and 90.70\% AUC. 
Note that our approach only accesses very few OOD samples during testing, and does not re-train or modify the trained OOD detection model. This result suggests a practical usage scenario, where a small number of accessible OOD samples can effectively reduce OOD FPR performance.

\begin{table*}[t]
\addtolength{\tabcolsep}{-1pt}
\caption{Evaluation of our OOD data enrolling approach using 10 samples in CIFAR-100 compared to the outlier exposure strategy using an auxiliary OOD dataset with around 80 million images. Our approach does not require model retraining, while the outlier exposure approach uses a huge auxiliary dataset for model training (see $\S$~\ref{sssec:improve_enroll}).
\vspace{-2mm}
}
\label{tab:outlier exposure}
\centerline{
\setlength{\tabcolsep}{0.1mm}
\begin{tabular}{l|cc|cc|cc|cc|cc|cc}
\toprule
 & \multicolumn{2}{c}{SVHN} & \multicolumn{2}{c}{Places365} & \multicolumn{2}{c}{LSUN} & \multicolumn{2}{c}{iSUN} & \multicolumn{2}{c}{Texture} & \multicolumn{2}{c}{Average} \\
\midrule
Model            & FPR$_{95}${\footnotesize $\downarrow$} & AUC{\footnotesize $\uparrow$}   & FPR$_{95}${\footnotesize $\downarrow$} & AUC{\footnotesize $\uparrow$} & FPR$_{95}${\footnotesize $\downarrow$} & AUC{\footnotesize $\uparrow$}    & FPR$_{95}${\footnotesize $\downarrow$} & AUC{\footnotesize $\uparrow$}    & FPR$_{95}${\footnotesize $\downarrow$} & AUC{\footnotesize $\uparrow$}    & FPR$_{95}${\footnotesize $\downarrow$} & AUC{\footnotesize $\uparrow$}   \\
\midrule
MMEL             & 9.70         & 97.24      & 58.72         & 81.01         & 5.12        & 98.54      & 34.57       & 88.52      & 44.31        & 86.42        & 30.48         & 90.70      \\
ICE              & 22.41       & 94.71      & 49.00         & 87.55         & 25.37       & 94.15      & 39.05       & 88.45      & 38.95        & 89.68        & 34.96         & 90.90 \\
\bottomrule
\end{tabular}
}
\vspace{-1mm}
\end{table*}

\begin{table*}[t]
\caption{The test-time OOD detection results using ImageNet-100 as the ID dataset. Three enrollment scenarios include enrolling OOD samples, novel-class samples, and both (see $\S$\ref{sssec:enroll_class}). We regard the rest classes in ImageNet as novel ID classes for enrollment. 
\vspace{-2mm}
}
\label{tab:enroll_imagenet}
\centerline{
\setlength{\tabcolsep}{-0.1mm}
\begin{tabular}{l|cc|cc|cc|cc|cc}
    \toprule
    & \multicolumn{2}{c}{SUN} & \multicolumn{2}{c}{Place365} & \multicolumn{2}{c}{Textures} & \multicolumn{2}{c}{iNaturalist} & \multicolumn{2}{c}{Average} \\
    \midrule
    & FPR$_{95}$$\downarrow$      & AUC$\uparrow$        & FPR$_{95}$$\downarrow$         & AUC$\uparrow$           & FPR$_{95}$$\downarrow$      & AUC$\uparrow$        & FPR$_{95}$$\downarrow$          & AUC$\uparrow$            & FPR$_{95}$$\downarrow$        & AUC$\uparrow$          \\
\midrule
No enrollment & 63.50  & 75.84 & 69.90  & 73.17 & 53.88 & 81.69 & 69.40  & 76.39 & 64.17 & 76.77 \\
OOD enrollment                    & 50.41 & 81.85 & 58.21 & 78.11 & 50.14 & 82.96 & 31.64 & 91.11 & 47.60  & 83.51 \\
Class enrollment    & 60.44 & 78.41 & 67.52 & 75.23 & 46.29 & 86.36 & 60.60  & 81.65 & 58.71 & 80.41 \\
\makecell{Class + OOD enrollment} & \textbf{46.78} & \textbf{83.71} & \textbf{54.39} & \textbf{80.91} & \textbf{41.93} & \textbf{88.15} & \textbf{29.92} & \textbf{91.99} & \textbf{43.25} & \textbf{86.19} \\
    \bottomrule
\end{tabular}
}
\vspace{-3mm}
\end{table*}

\vspace{-2mm}
\subsubsection{Effects of enrolling novel classes}
\label{sssec:enroll_class}
\vspace{-1mm}

We further investigate the enrollment properties by introducing novel classes as part of the ID data during test time. This experiment aims to explore the possibility of expanding the ID space without necessitating model retaining. Specifically, using ImageNet-100 as the ID dataset, we enroll the additional 900 classes from ImageNet without model retraining. The OOD detection is then conducted using the same settings outlined in $\S$\ref{sssec:improve_enroll} on four OOD evaluation datasets.

Given that the novel classes are unseen to the trained model, we employ the steps described in $\S$~\ref{ssec:score_calculation} to extract embeddings for the observed new-class samples. These embeddings are aggregated to form a positive sample prototype for distance measurement, serving as a score $S_{Novel}(\mathbf{z})$ for the novelty class. This term is subsequently added in the final scoring calculation: $S(\mathbf{z})-S_{OOD}(\mathbf{z})+S_{Novel}(\mathbf{z})$. Samples in close proximity to the enrolled OOD samples yield high $S_{OOD}(\mathbf{z})$, while those near the novel classes obtain high $S_{Novel}(\mathbf{z})$. This approach allows us to observe the effects of enrolling different types of samples, particularly in scenarios involving novel classes.

Table~\ref{tab:enroll_imagenet} presents the OOD detection results under various scenarios of test-time enrollment. In the `Class enrollment' scenario, we enroll only one sample for each novel class, while in `OOD enrollment', we enroll 10 OOD samples. The results indicate a significant 16.57\% FPR drop and 6.74\% AUC increase when employing `OOD enrollment', compared to direct OOD detection without any sample enrollment across all 1,000 classes in the ImageNet dataset. Even enrolling just one sample for each novel class results in a 5.46\% FPR reduction. The optimal performance is achieved by simultaneously enrolling both the novel classes and 10 OOD samples, yielding a remarkable 43.25\% FPR$_{95}$ and 86.19\% AUC. We ascribe the generalization potentials of MMEL to the increased manifolds that enable adaptively adjust cluster spaces either for novel ID classes or OOD examples.


\section{Conclusion}
\label{sec:conclusion}

The detection of out-of-distribution (OOD) instances is crucial for the safe and reliable deployment of AI in real-world scenarios. Traditional OOD detection research has ignored the data diversity in embedding learning and suffered the distortation risk in modeling the whole ID data in a single manifold structure.
In this work, we introduce a novel multi-manifold embedding learning (MMEL) framework that incorporates hypersphere and hyperbolic embeddings, coupled with a prototype-aware KNN scoring function, to enhance the robustness of in-distribution (ID) representations. Our proposed framework demonstrates significant performance boost. With flexibility of modeling multi-manifold data, we put forth an OOD sample enrollment scenario to further diminish FPR for real-world applications. Further experiments highlight the potential to enroll either ID or OOD samples with minimal samples collected during test time.


For future work, exploring manifold optimization for ID data preservation and extending the MMEL for continual OOD detection with manifold adaptation can substantially enhance usability of OOD detection.





%
%
\bibliographystyle{splncs04}
\bibliography{main}
\end{document}